%% file: main.tex
\documentclass{article} % For LaTeX2e

\usepackage{makecell}
\usepackage{multirow}
\usepackage{amsmath}
\newcommand{\smlpm}{\scriptstyle\pm}

\PassOptionsToPackage{round}{natbib}

\usepackage{paper}

\usepackage[utf8]{inputenc} % allow utf-8 input
\usepackage[T1]{fontenc}    % use 8-bit T1 fonts
\usepackage{hyperref}       % hyperlinks
\usepackage{url}            % simple URL typesetting
\usepackage{booktabs}       % professional-quality tables
\usepackage{amsfonts}       % blackboard math symbols
\usepackage{nicefrac}       % compact symbols for 1/2, etc.
\usepackage{microtype}      % microtypography
\usepackage{xcolor}         % colors

\usepackage{graphicx}
\usepackage{subcaption}
\usepackage{algorithm}
\usepackage{algorithmic}
\usepackage{caption}
\usepackage{wrapfig}
\usepackage{dsfont}

\definecolor{darkblue}{RGB}{0,0,139}
\definecolor{darkgreen}{RGB}{0,100,0}

\hypersetup{
    colorlinks=true,
    linkcolor=darkblue,
    citecolor=darkblue,
    urlcolor=darkgreen
}

\newcommand{\smartref}[1]{\hyperref[#1]{\ref{#1}}}

\title{Learning Transferable Skills using \\ Goal-Conditioned Bisimulation}

\author{%
  Mohammad Amin Abbasfar \thanks{Corresponding author} \quad Farbod Azimmohseni \quad Mohammad Hossein Rohban \\
  Department of Computer Engineering \\
  Sharif University of Technology \\
  \texttt{\{amin.abbasfar01, far.azimmohseni.82, rohban\}@sharif.edu} \\
}

\begin{document}

\maketitle

\setcounter{footnote}{0}

\input{tex/0_abstract}

\section{Introduction}
\input{tex/1_introduction}

\section{Related Work}
\label{related_work}
\input{tex/2_related_work}

\section{Preliminaries}
\label{preliminaries}
\input{tex/3_preliminaries}

\section{Learning Transferable Skills using Bisimulation}
\input{tex/4_method}

\section{Experiments}
\input{tex/5_experiments}

\section{Conclusion}
\input{tex/6_conclusion}

\newpage

\bibliographystyle{plainnat}
\bibliography{references}

\newpage
\appendix

\section{Additional Visualizations}
\label{appendix:viz}
\input{tex/appendix/additional_visualizations}

\section{Additional Experiments}
\label{appendix:add_exp}
\input{tex/appendix/additional_experiments}

\section{Additional Ablation Studies}
\label{appendix:add_ablation}
\input{tex/appendix/additional_ablations}

\section{Environment Details}
\label{appendix:env}
\input{tex/appendix/environment_details}

\section{Implementation Details}
\label{appendix:impl}
\input{tex/appendix/implementation_details}

\newpage

\end{document}

%% file: tex/0_abstract.tex
\begin{abstract}
    Unsupervised skill discovery has emerged as a promising approach for leveraging reward-free datasets to pretrain general-purpose policies. However, current skill discovery methods either require access to expert data or exhibit limited generalization, failing to transfer effectively to previously unseen layouts. A key challenge is to learn representations that capture the temporal structure of the environment while remaining robust to variations across layouts. To address this issue, we present an objective for learning \emph{action-aware temporal representations} that satisfy the functional equivariance property while preserving the local temporal structure of the environment. Building upon this embedding, we further propose \emph{unsupervised skill discovery using bisimulation}, which learns transferable skills by conditioning the behavior of skills exclusively on the subset of state features that directly affect their execution. This enforces invariant behavior across different layouts, enabling skills to transfer effectively to other configurations. Finally, through comprehensive empirical evaluations, we show that skills learned in a given environment can be effectively applied to solve downstream tasks in various environment layouts, demonstrating strong out-of-distribution generalization. \footnote{The code and videos are available at \url{https://sites.google.com/view/usb-atr/}.}
\end{abstract}

%% file: tex/1_introduction.tex
Leveraging large unlabeled datasets has substantially contributed to the advancement of deep learning in recent years. Unsupervised and self-supervised techniques such as masked language modeling in natural language processing \citep{devlin2019bert} and contrastive losses in computer vision \citep{chen2020simple} have enabled deep neural networks to benefit from vast and low-cost unlabeled data. Numerous attempts have been made to apply the same strategy in the context of reinforcement learning (RL), where the goal is to pretrain the agent's policy using only pre-collected offline datasets \citep{chebotar2021actionable,reed2022gato,park2023hiql,park2024hilp,openx2024rtx}.

Unsupervised skill discovery has been recognized as a promising approach for pretraining such policies. These methods aim to learn a versatile set of latent‑conditioned policies, each demonstrating a distinct behavior, such that together they cover the state space. Prior work has proposed various objectives to train such skill policies, either via behavioral cloning \citep{pertsch2021spirl,ajay2021opal} or by diversifying their state visitation distribution \citep{eysenbach2019diayn,park2023csd,park2024hilp}. These skills can then be used to perform downstream tasks in a zero-shot manner, or harnessed as low-level policies within a hierarchical approach \citep{eysenbach2019diayn,sharma2020dads}.

Many applications of reinforcement learning in locomotion and robotic manipulation involve frequently occurring \emph{temporally-extended actions} that appear during the accomplishment of a wide range of tasks \citep{hauskrecht1998hierarchical,sutton1999temporal,pateria2021hrl}.
For instance, in the AntMaze environment \citep{fu2020d4rl,park2025ogbench}, where a quadruped agent must navigate to different goals within a maze, the ability to move in the four cardinal directions constitutes a sufficient set of skills for performing any task, regardless of maze layout. Similarly, in the Kitchen environment \citep{gupta2019kitchen}, which consists of several long-horizon manipulation tasks, a temporally-extended action such as placing the kettle from the cupboard onto the oven top should be performed independently of other environmental features, such as whether the microwave door is open or closed. In such structured environments, learning reusable skills is essential for generalizing to previously unseen configurations. 

Despite the importance of such reusable behaviors in structured domains, existing skill discovery methods are not explicitly designed to account for this structure. For instance, the Hilbert foundation policy (HILP) \citep{park2024hilp}, which is a state-of-the-art distance-maximizing skill discovery method, learns skills that span the state space reasonably well. However, it struggles to generalize to out-of-distribution layouts \citep{sobal2025pldm}. To illustrate, suppose HILP learns to move a kettle only when the microwave door is closed. In a scenario where the microwave door is open, despite being irrelevant to the task, the behavior of the kettle-moving skill may change drastically, making it incapable of completing the task. In contrast to distance-maximizing skill discovery methods, behavioral cloning-based methods offer better out-of-distribution generalization \citep{pertsch2021spirl}. Nevertheless, they fall short when using suboptimal datasets, as they tend to replicate the exact action sequences that exist in the dataset.

\begin{figure}[t]
    \centering
    \begin{subfigure}[b]{0.45\textwidth}
        \centering
        \includegraphics[width=\textwidth]{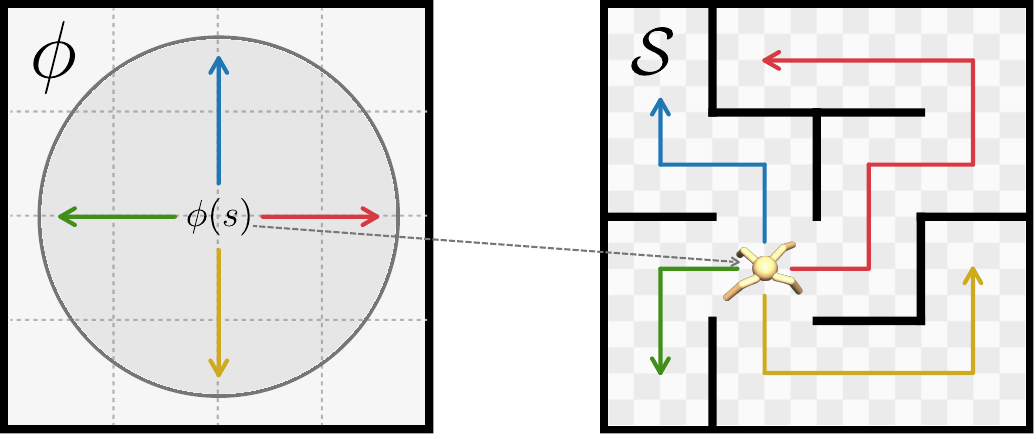}
        \caption{Skills discovered by HILP}
        \label{fig:fig1a}
    \end{subfigure}
    \hfill
    \begin{subfigure}[b]{0.45\textwidth}
        \centering
        \includegraphics[width=\textwidth]{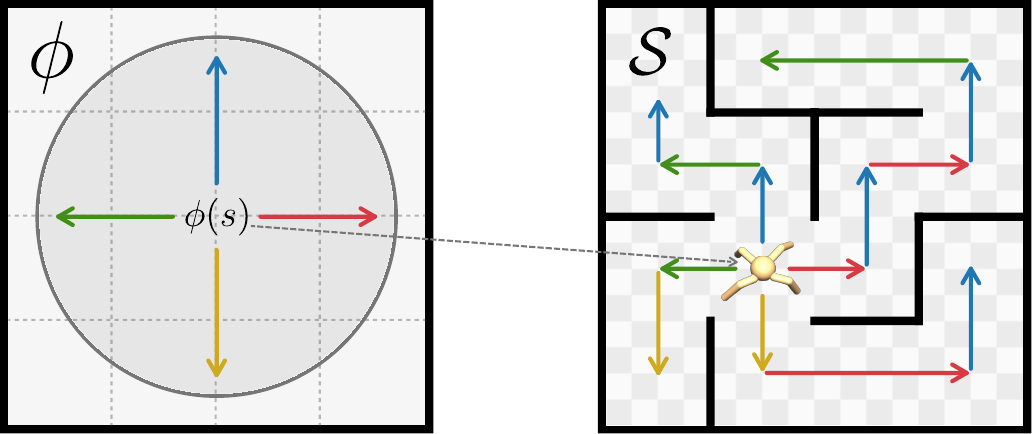}
        \caption{Skills discovered by USB+ATR (ours)}
        \label{fig:fig1b}
    \end{subfigure}
    \caption{\emph{(Left)} HILP's skills overfit to the training layout despite spanning the state space, limiting their ability to generalize to new layouts. \emph{(Right)} USB+ATR extracts more fine-grained and layout-agnostic skills, which transfer effectively to unseen layouts.\protect\footnotemark}

    \label{fig:fig1}
\end{figure}
\footnotetext{This figure is adapted from \citet{park2024hilp}.} 

Motivated by this, we aim to extract skills that are transferable to out-of-distribution layouts, even when only \emph{a single} training layout and low-quality datasets are available during training. Our method consists of two components. First, we introduce an objective for training an \emph{action-aware temporal representation} (ATR) that aims to learn a structured state embedding with \emph{functional equivariance} \citep{hansen-estruch2022gcb} property, while preserving local temporal distances. This allows us to enforce each skill to show \emph{monotonic behavior} while retaining the benefits of temporal representations when learning from suboptimal data. To train such representations, we employ bisimulation metrics \citep{ferns2014bisimulation}, which have demonstrated well-established capabilities for improving generalization \citep{zhang2021dbc,agarwal2021psm}. In particular, we use \emph{goal-conditioned bisimulation} \citep{hansen-estruch2022gcb} to enforce this property for state-goal pairs that share similar underlying action sequences.

Second, we build upon this representation to propose \emph{unsupervised skill discovery using bisimulation} (USB), a modification of HILP in which the behavior of skills is conditioned only on the state features that are relevant to their execution, rather than on the full state. To achieve this, we again employ bisimulation metrics to learn a state representation in which states with similar transition dynamics are embedded close to one another. Conditioning each skill on this representation encourages the learned skills to exhibit consistent behavior across dynamically similar states, even when the corresponding action sequences are not present in the dataset for all such states. In this way, we encourage skills to disregard layout-specific features of the environment and to acquire generic behaviors that improve their transferability. These generic skills can then be equipped with a planning approach or composed by a high-level controller to accomplish downstream tasks across diverse layouts.

Finally, to evaluate the out-of-distribution generalization capability of our method, we assess the learned skills in both a test-time planning setup and a hierarchical approach. By conducting experiments across environments with different layout maps, we demonstrate that our method offers strong generalization and outperforms existing state-of-the-art offline goal-conditioned RL methods.

In summary, our main contributions in this work are as follows: 
1) We propose ATR, a novel structured temporal representation for the states with the functional equivariance property.
2) Building on this representation, we propose USB, an unsupervised skill discovery method for pretraining latent-conditioned policies that learn generic and transferable skills.
3) We extend the popular OGBench \citep{park2025ogbench}, a standard offline goal-conditioned RL benchmark, by adding a feature for generating random maze layouts and datasets, enabling a more thorough evaluation of the generalization ability of different methods.
4) Through experiments in several maze environments, we empirically demonstrate that USB+ATR learns effective and transferable skills across different layouts and offers strong out-of-distribution generalization.

%% file: tex/2_related_work.tex
\textbf{Offline goal-conditioned RL.}
Goal-conditioned objectives have been widely used in unsupervised RL to learn latent representations for states that preserve temporal distances (i.e., the minimum number of steps required to reach a goal from a given state under an optimal policy). 
Under certain assumptions, the optimal goal‑conditioned value function is equivalent to the discounted temporal distance between pairs of states \citep{kaelbling1993gcloss}.
Inspired by this, prior work has explored learning universal value functions as a proxy for learning temporal representations
\citep{ghosh2023passive,park2023hiql}.
Another line of work has sought to learn such representations by incorporating contrastive losses \citep{eysenbach2022crl,ma2023vip,myers2024cmd}.
Since optimal goal-conditioned value functions satisfy the triangle inequality \citep{liu2023mrn,pitis2020inductive}, 
several model architectures \citep{wang2022iqe,liu2023mrn} and objective functions \citep{wang2023qrl,myers2025tmd} have been proposed to leverage this inductive bias in order to learn temporal embeddings more efficiently.

\textbf{Unsupervised skill discovery.}
Skill discovery is a powerful approach in unsupervised RL aiming to pretrain a latent-conditioned policy that encodes a diverse set of behaviors. These skills are often trained by behavioral cloning \citep{pertsch2021spirl,ajay2021opal}, maximizing mutual information between skills and their future state distributions \citep{gregor2017vic,eysenbach2019diayn,sharma2020dads}, or maximizing the distance between these distributions \citep{park2022lsd,park2023csd,park2024metra,park2024hilp}. 
The closest to our work is HILP \citep{park2024hilp}, which incorporates a temporal representation to train each skill by encouraging movement along a random unit vector in that representation. This is achieved by defining directional rewards for each skill, computed as the inner product between the temporal displacement caused by executing this skill and its corresponding target vector.

\textbf{Bisimulation.}
Many approaches have been proposed for clustering states in general Markov decision processes (MDPs) to make a form of state abstraction. Bisimulation \citep{larsen1991bisimulation} is the strictest approach, which groups only states that are indistinguishable and behaviorally equivalent under the optimal value function.
Bisimulation metrics \citep{ferns2011bisimulation,ferns2014bisimulation} relax this assumption by defining a notion of behavioral similarity between states. \citet{castro2020scalable} introduces on-policy bisimulation, which is a scalable and computationally efficient approach for approximating bisimulation metrics by using the similarity between on-policy behaviors of states, for deterministic MDPs. \citet{zhang2021dbc} propose a gradient-based method for learning a latent representation for high-dimensional state spaces in general MDPs with stochastic transitions. 
Finally, our action-aware temporal representation is based on the goal-conditioned extension of bisimulation metrics \citep{hansen-estruch2022gcb}, where an embedding of state-goal pairs is learned to cluster such pairs with similar underlying reward or action sequences.

%% file: tex/3_preliminaries.tex
\textbf{Problem setting.} 
We assume our environment is an MDP defined by $(\mathcal{S}, \mathcal{A}, \mathcal{P}, \mathcal{R}, \gamma)$, where $\mathcal{S}$ is the state space, $\mathcal{A}$ is the action space, $\mathcal{P}: \mathcal{S} \times \mathcal{A} \rightarrow \Delta (\mathcal{S})$ is the transition dynamics, and $\mathcal{R}: \mathcal{S} \rightarrow \mathbb{R}$ is the reward function. Following prior works in offline goal-conditioned RL \mbox{\citep{park2023hiql,wang2023qrl}}, we also assume that the MDP is deterministic. We are also given an offline unlabeled dataset $\mathcal{D}$ containing trajectories $\tau=(s_0, a_0, s_1, \dots, s_T )$ collected from an unknown behavior policy $\pi^b$.

We want to learn a skill-conditioned policy $\pi^l(a |s, z)$, where $z\in \mathcal{Z}$ denotes the latent embedding of skills. We aim for the policy $\pi^l$ to learn a diverse range of behaviors for spanning the state space, while also being transferable to previously unseen layouts.

\textbf{Representation learning using bisimulation metrics.} 
On-policy bisimulation metrics can be expressed in the following recursive formulation \citep{castro2020scalable}:
\begin{equation}
\label{eq:bisim_pi}
    d(s_i; s_j) = \mathbb{E}_{a_i \sim \pi(\cdot | s_i), a_j \sim \pi(\cdot | s_j)}\Big[|\mathcal{R}(s_i) - \mathcal{R}(s_j)| + \gamma \mathcal{W}_1(\mathcal{P}(\cdot | s_i, a_i), \mathcal{P}(\cdot | s_j, a_j); d)\Big],
\end{equation}
where $\mathcal{W}_1$ denotes the 1-Wasserstein distance.
As proposed by \citet{zhang2021dbc}, replacing the 1‑Wasserstein distance with the 2‑Wasserstein distance allows this recursive formulation to be used as an objective for training a latent representation $h$. We adopt this objective for our offline setting:
\begin{equation}
\label{eq:bisim_loss}
    \mathcal{L}_h = \Big( \| h(s_i) - h(s_j) \|_1 
    - |\mathcal{R}(s_i) - \mathcal{R}(s_j)| - \gamma \mathcal{W}_2(\mathcal{P}(\cdot | \bar{h}(s_i), a_i), \mathcal{P}(\cdot | \bar{h}(s_j), a_j)) \Big)^2,
\end{equation}
where $\bar{\cdot}$ denotes a stop-gradient. Under the deterministic MDP assumption, the 2‑Wasserstein distance term simplifies to:
$\mathcal{W}_2(\mathcal{P}(\cdot | \bar{h}(s_i), a_i), \mathcal{P}(\cdot | \bar{h}(s_j), a_j))$ = $\| \bar{h}(s_i') - \bar{h}(s_j') \|_2$.

%% file: tex/4_method.tex
\begin{figure}[t]
    \centering
    \begin{subfigure}[b]{0.22\textwidth}
        \centering
        \includegraphics[width=\textwidth]{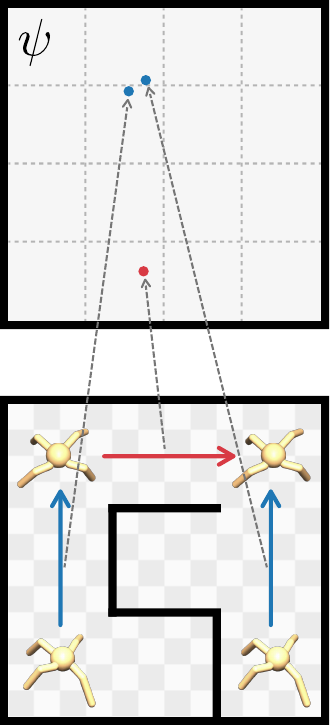}
        \caption{}
        \label{fig:fig2a}
    \end{subfigure}
    \hfill
    \begin{subfigure}[b]{0.22\textwidth}
        \centering
        \includegraphics[width=\textwidth]{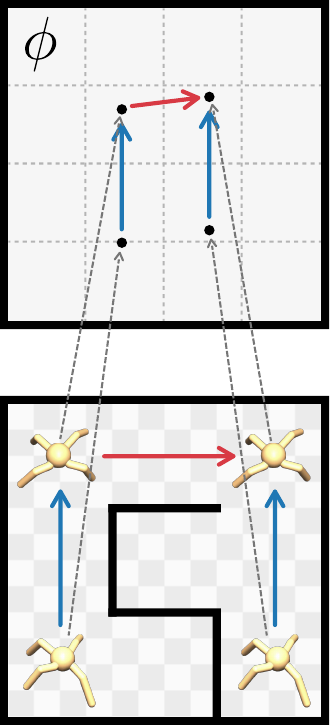}
        \caption{}
        \label{fig:fig2b}
    \end{subfigure}
    \hfill
    \begin{subfigure}[b]{0.22\textwidth}
        \centering
        \includegraphics[width=\textwidth]{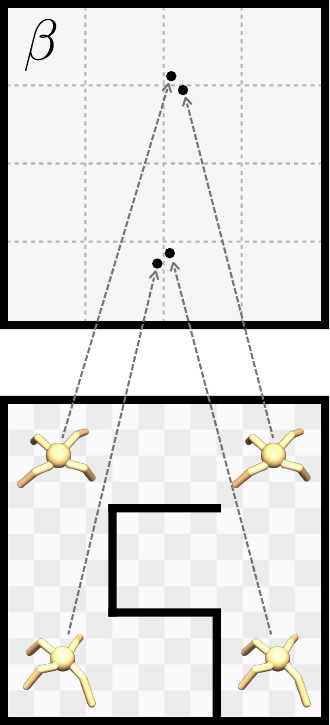}
        \caption{}
        \label{fig:fig2c}
    \end{subfigure}
    \hfill
    \begin{subfigure}[b]{0.22\textwidth}
        \centering
        \includegraphics[width=\textwidth]{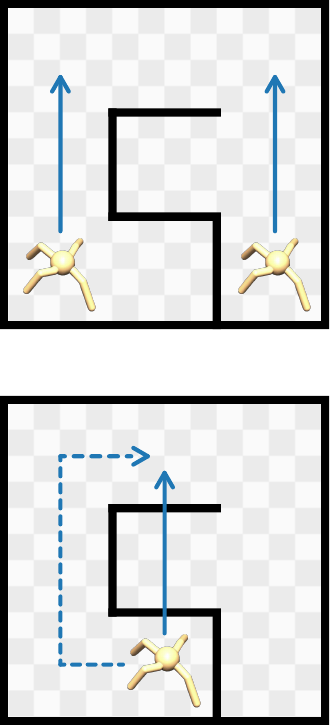}
        \caption{}
        \label{fig:fig2d}
    \end{subfigure}
    \caption{(a) We first train a state-goal pair encoder $\psi$ using goal-conditioned bisimulation, which encodes pairs with similar underlying action sequences close to each other. (b) Next, we train an action-aware temporal representation $\phi$ that preserves local temporal dependencies while respecting functional equivariance enforced by $\psi$. (c) Based on the similarity between the local structures shaped in representation $\phi$, we train a representation $\beta$ using bisimulation to encode dynamically similar states to spatially close latent vectors. (d) Lastly, we train the skill policy using HILP directional state-spanning rewards by conditioning these skills on the $\beta$ latents. Unlike HILPs, which exploit the layout structure to greedily maximize the reward (dashed line), USB instead learns monotonic behaviors across dynamically similar states, making the resulting skills transferable to unseen layouts (solid line).}
    \label{fig:fig2}
\end{figure}

\subsection{Motivation}
Current unsupervised skill discovery methods that are based on temporal representations, such as HILP, tend to generalize poorly to unseen layouts \citep{sobal2025pldm}. In the absence of structural constraints, temporal embeddings are shaped by the specific layout of the environment and its corresponding temporal distances. 
Because the temporal relationships among states vary across layouts, skill discovery methods that rely solely on covering the temporal structure of a particular layout often fail to transfer to new environment configurations.
To achieve better transferability, it is necessary to learn more primitive yet frequently occurring action sequences that remain invariant across different layouts (Figure \smartref{fig:fig1}).

Inspired by this observation, we argue that a generalizable skill discovery method must satisfy two desiderata. First, each skill should be capable of executing one or more temporally extended actions, and collectively cover the entire state space. Second, the behavior of each skill should be conditioned only on those state features that are necessary for its execution. The first property ensures the versatility and effectiveness of the learned skills within the training layout, while the second enforces behavioral consistency across dynamically similar states, thereby enabling transfer even to previously unseen layouts.

\subsection{Action-Aware Temporal Representation}
\label{sec:atr}

To train a skill policy with the properties described in the previous section, we first learn an action-aware temporal representation that preserves local temporal relationships while respecting functional equivariance. By definition, a universal temporal representation must (1) maximally spread out states while (2) preserving their local temporal distance constraints \citep{wang2023qrl}. However, freely maximizing the distances between states can eventually lead the representation to overfit to the specific layout, as the representation is encouraged to spread states according to layout-dependent features. Therefore, rather than freely maximizing distances between states, we aim to spread states while simultaneously enforcing functional equivariance when learning temporal distances. Although this design sacrifices some properties of a temporal representation, as our representation is no longer a universal temporal representation, it provides the additional benefit of generalizing across different layouts.

To enforce functional equivariance in the representation, we adopt goal-conditioned bisimulation  \citep{hansen-estruch2022gcb}. Goal-conditioned bisimulation extends the classical notion of bisimulation metrics to goal-conditioned settings by grouping together state-goal pairs that correspond to analogous tasks. 
Inspired by \citet{agarwal2021psm}, and in contrast to the typical practice in the bisimulation literature, we define our bisimilarity metric as the $\ell_1$ distance between actions rather than rewards. However, unlike \citet{agarwal2021psm}, who measure distances using optimal actions, we sample actions from our dataset for which we do not have optimality assumptions. Instead, the goal-conditioned formulation allows us to exploit such potentially suboptimal actions, since similarity is defined with respect to achieving the specified goals rather than relying on optimal behavior.

Following the approach used in \citet{hansen-estruch2022gcb} and similar to Equation (\smartref{eq:bisim_loss}), we define an embedding $\psi(s, g)$ 
\footnote{For consistency with the notation used in other skill discovery methods, we reserve 
$\phi$ for the final temporal representation and 
$\psi$ for the goal-conditioned embedding, even though this differs from the notation in \citet{hansen-estruch2022gcb}.}
such that the $\ell_1$ distance between embeddings of state-goal pairs reflects their bisimilarity.
% as specified in Equation (\smartref{eq:gcb_pi}). 
Accordingly, the loss function for learning the representation $\psi$ is given by:
\begin{equation}
\label{eq:gcb_loss}
    \mathcal{L}_\psi = \Big( \| \psi(s_i, g_i) - \psi(s_j, g_j) \|_1 - \|a_i - a_j\|_1 - \gamma_\psi \| \bar{\psi}(s_i', g_i) - \bar{\psi}(s_j', g_j) \|_2 \Big)^2,
\end{equation}
where $(s_i, a_i, s_i', g_i)$ and $(s_j, a_j, s_j', g_j)$ denote state, next state, action, and goal tuples sampled from the dataset $\mathcal{D}$.
As a result of this objective, $\psi$ embeds bisimilar state-goal pairs close to one another in the representation space (Figure \smartref{fig:fig2a}).
Note that this representation captures similarities between state-goal pairs based on their associated action sequences and does not encode any notion of temporal distance. Indeed, the same action sequence can correspond to different temporal distances when executed from different initial states. 
Therefore, we use $\psi$ only to spread states across the temporal representation in a structured manner.

We can now define our action-aware temporal representation 
$\phi$ as:
\begin{equation}
\label{eq:phi_def}
    \phi(s_i) - \phi(s_i') := \psi(s_i, s_i') - \mathbb{E}_{s \in \mathcal{S}}[\psi(s, s)],
\end{equation}
in which $\mathbb{E}_{s \in \mathcal{S}}[\psi(s, s)]$ acts as a normalization term to maximally spread the action sequences in different directions. The choice of the normalizer was determined empirically, and an ablation study examining different choices of the normalizer is provided in Appendix \smartref{appendix:add_ablation}.

Using this objective, we encourage ATR to capture the encoded directions learned by $\psi$, while the constraints imposed by the metric space ensure that it satisfies the triangle inequality (Figure \smartref{fig:fig2b}). 
For learning such a representation, the resulting objective function is defined as:
\begin{equation}
\label{eq:atr_loss}
    \mathcal{L}_\phi=\left\|(\phi(s_i) - \phi(s'_i)) - (\bar{\psi}(s_i, s'_i) - \mathbb{E}_{s \sim \mathcal{D}}[\bar{\psi}(s, s)]) \right\|_2^2,
\end{equation}
% where $s_i$ and $s_i'$ are consecutive states sampled from $\mathcal{D}$.

As a consequence of enforcing the triangle inequality on the representation of states, the displacement in temporal representation caused by a given action sequence is no longer identical across all states but is instead influenced by the underlying dynamics of the surrounding states as well. This property allows us to identify similarities in state dynamics by comparing the temporal differences produced by the same action sequence across states. We leverage this characteristic in Section \ref{sec:usb} to capture dynamically similar states. Additional visualizations of ATR along with comparisons to the temporal embeddings used by HILP are provided in Appendix \smartref{appendix:viz}.

We emphasize this point as particularly important. Although \citet{hansen-estruch2022gcb} learn a very similar embedding, our definition of $\phi$
differs in two key aspects. First, unlike \citet{hansen-estruch2022gcb}, who uses goal-conditioned rewards as a metric and therefore learns a temporal representation, we use actions as the metric. This choice fundamentally changes the properties of $\phi$
and provides a stronger notion of functional equivariance. Second, because $\psi$ does not provide temporal information and only the direction of vectors is informative in our case, we minimize $\mathcal{L}_\phi$ (Equation \smartref{eq:atr_loss}) only on adjacent states. The full embedding is then constructed from these local relationships rather than from distant state-goal pairs.

\begin{algorithm}[t]
\caption{Training Algorithm for USB+ATR}
\label{alg:training}
\begin{algorithmic}[1]
\WHILE{not converged}
    \STATE Sample $(s_i, a_i, s_i', g_i), (s_j, a_j, s_j', g_j) \sim \mathcal{D}$
    
    \STATE Update $\psi$ to minimize $\mathcal{L}_\psi((s_i, a_i, s_i', g_i), (s_j, a_j, s_j', g_j))$
    
    \STATE Update $\phi$ to minimize $\mathcal{L}_\phi((s_i, s_i'), (s_j, s_j'))$
    
    \STATE Sample $a_k \sim \mathcal{D}$
    \STATE Update $\beta$ to minimize $\mathcal{L}_\beta(s_i, s_j, a_k)$
    
    \STATE Update $f$ to minimize $\| f(s_i, a_i) - s_i' \|^2_2$

    \STATE Sample uniformly $z \sim \mathcal{Z}$ s.t. $\|z\|_2=1$
    \STATE Compute directional reward $\mathcal{R}(s_i, z, s_i')$
    
    \STATE Train $\pi^l(\cdot | \bar{\beta}(s_i), z)$ with any off-the-shelf offline RL algorithm
\ENDWHILE
\end{algorithmic}
\end{algorithm}

\subsection{Unsupervised Skill Discovery using Bisimulation}
\label{sec:usb}
Using ATRs instead of universal temporal representations reduces the variance in the action sequences encoded along each direction because of their functional equivariance property. As a result, when skills are trained with directional reward functions, they tend to exhibit more monotonic behavior. However, because each skill greedily maximizes its cumulative reward, it still may exploit the environment’s layout-specific structure by deviating from its target direction to obtain distant rewards. Such deviations conflict with the intended monotonic behavior of the skills.

To prevent skills from \emph{memorizing} the environment layout, we condition the skill policy only on the subset of state features that determine their dynamics, as these provide the information necessary for executing the skill. This constraint enforces skills to behave consistently across dynamically similar states and improves generalization to out-of-distribution states by reducing the influence of irrelevant state features.
To this end, once again we employ bisimulation metrics to learn a representation $\beta(s)$ in which states with similar transition dynamics are embedded spatially close to each other (Figure \smartref{fig:fig2c}). To define a metric that captures this relationship between states, we leverage the emergent characteristic of ATRs discussed in Section \smartref{sec:atr}. ATRs maintain functional equivariance as long as it does not disrupt the temporal dependencies among neighboring states. Therefore, any variation in how a specific action sequence affects the temporal embedding must arise from differences in the underlying state dynamics. Based on this property, we define our metric as the dissimilarity between the \emph{temporal displacement} resulting from executing the same action from two distinct states. Using this definition, the recursive formulation of bisimulation can be expressed as:
\begin{align}
    d(s_i; s_j) = \mathbb{E}_{a_k \sim \operatorname{Unif}(\mathcal{A}), s_i' \sim \mathcal{P}(\cdot | s_i, a_k), s_j' \sim \mathcal{P}(\cdot | s_j, a_k)} 
    \Big[& \| (\phi(s_i') - \phi(s_i)) - (\phi(s_j') - \phi(s_j)) \|_1 + \nonumber \\ 
    &\gamma \mathcal{W}_2(\mathcal{P}(\cdot | s_i, a_k), \mathcal{P}(\cdot | s_j, a_k); d) \Big],
\end{align}
where \(\operatorname{Unif}(\mathcal{A})\) denotes the uniform distribution over the action space \(\mathcal{A}\).

Since our metric measures the difference between the temporal displacement produced by applying \emph{the same} action $a_k$ to both $s_i$ and $s_j$, converting the on-policy formulation to the offline setting is challenging, as such matching actions may not be available in the dataset. To address this issue, we first learn a dynamics model $f$ that predicts the next state. We then construct matching action pairs by sampling two states and a shared action, and generating their corresponding next states using $f$. The final loss for $\beta(s)$ is then defined as:
\begin{align}
    \mathcal{L}_\beta= \Big(\|\beta(s_i) - \beta(s_j)\|_1 - 
    \| [\bar{\phi}(\bar{f}(s_i, a_k)) - \bar{\phi}(s_i)] &- [\bar{\phi}(\bar{f}(s_j, a_k)) - \bar{\phi}(s_j)] \|_1 - \nonumber
    \\& \gamma_\beta || \bar{\beta}(\bar{f}(s_i, a_k)) - \bar{\beta}(\bar{f}(s_j, a_k)) ||_2 \Big)^2.
\end{align}

Lastly, we train the skill policy $\pi^l(\cdot | \bar{\beta}(s), z)$ which is conditioned on the $\beta$-representation of states. Following HILP, we uniformly sample skill embeddings from $\{z \in \mathcal{Z}: \| z\|_2 = 1\}$, and train the latent-conditioned policy using the following directional reward:
\begin{equation}
    \mathcal{R}(s, z, s')=\langle \phi(s') - \phi(s), z \rangle,
\end{equation}
where $\langle \cdot, \cdot \rangle$ denotes the inner product between vectors. By training USB on ATR, each skill acquires monotonic and robust behaviors, enabling transferability to previously unseen layouts (Figure \smartref{fig:fig2d}). Bringing all components together, the complete method is summarized in Algorithm \smartref{alg:training}.

\input{tables/hierarchical_layout_generalization}

%% file: tables/hierarchical_layout_generalization.tex
\begin{table}[t]
\centering
\caption{\textbf{Evaluation of transferability to new layouts with a high-level controller.} The average success rates with standard errors are calculated across 3 layouts and 5 seeds for each (15 runs in total).}
\label{table:layout_gen}
\fontsize{9pt}{10pt}\selectfont
\begin{tabular}{l l l | c c c c}
\toprule
\multicolumn{3}{c|}{Task} &  \multicolumn{4}{c}{Method}  \\
Environment     & Dataset  & Size  & Flat & OPAL & HILP & \textbf{USB+ATR} \\

\midrule

\multirow{4}{*}{\texttt{pointmaze}}
& \multirow{2}{*}{\texttt{navigate}} 
    & \texttt{medium} & $0.0 \smlpm 0.0$ & $0.0 \smlpm 0.0$ & $40.0 \smlpm 7.4$ & $\mathbf{91.3 \smlpm 4.2}$ \\
    && \texttt{large} & $0.0 \smlpm 0.0$ & $30.7 \smlpm 6.9$ & $44.4 \smlpm 7.5$ & $\mathbf{100.0 \smlpm 0.0}$  \\
\cmidrule(lr){2-7}
& \multirow{2}{*}{\texttt{stitch}}   
    & \texttt{medium} & $0.0 \smlpm 0.0$ & $2.1 \smlpm 2.1$ & $11.1 \smlpm 4.7$ & $\mathbf{75.6 \smlpm 6.5}$  \\
    && \texttt{large} & $0.0 \smlpm 0.0$ & $20.0 \smlpm 6.0$ & $28.9 \smlpm 6.8$ & $\mathbf{81.9 \smlpm 5.8}$  \\

\midrule
\multirow{6}{*}{\texttt{antmaze}}
    & \multirow{2}{*}{\texttt{navigate}} 
        & \texttt{medium} & $2.1 \smlpm 0.7$ & $\mathbf{99.0 \smlpm 0.2}$ & $57.7 \smlpm 6.3$ & ${92.3 \smlpm 1.9}$  \\
        && \texttt{large} & $0.0 \smlpm 0.0$ & $\mathbf{89.4 \smlpm 4.2}$ & $44.9 \smlpm 5.4$ & ${83.9 \smlpm 2.9}$  \\
    \cmidrule(lr){2-7}
    & \multirow{2}{*}{\texttt{stitch}}   
        & \texttt{medium} & $2.1 \smlpm 0.7$ & $\mathbf{99.2 \smlpm 0.2}$ & $59.0 \smlpm 6.8$ & $88.3 \smlpm 1.0$  \\
        && \texttt{large} & $0.0 \smlpm 0.0$ & $\mathbf{98.0 \smlpm 0.4}$ & $65.2 \smlpm 6.1$ & $78.9 \smlpm 2.2$  \\
    \cmidrule(lr){2-7}
    & \multirow{2}{*}{\texttt{explore}}   
        & \texttt{medium} & $2.1 \smlpm 0.7$ & $2.1 \smlpm 2.1$ & ${78.8 \smlpm 5.9}$ & $\mathbf{84.6 \smlpm 5.0}$  \\
        && \texttt{large} & $0.0 \smlpm 0.0$ & $0.0 \smlpm 0.0$ & $\mathbf{93.3 \smlpm 3.1}$ & $51.7 \smlpm 7.2$  \\

\midrule
\multicolumn{3}{l|}{\textbf{Average}} & $0.63$ & $44.05$ & $52.33$ & $\mathbf{82.85}$ \\

\bottomrule
\end{tabular}
\end{table}

%% file: tex/5_experiments.tex
\input{tables/planning_layout_generalization}

\subsection{Experimental Setup}
We conduct our experiments on OGBench \citep{park2025ogbench}, a recently proposed benchmark for offline goal-conditioned RL. In particular, we evaluate our method on two maze environments: \texttt{pointmaze} and \texttt{antmaze}, where the point and ant agents must navigate to a random goal location in mazes with different sizes. To study the sensitivity of each method to dataset quality, we use three dataset types provided by OGBench: \texttt{navigate}, \texttt{stitch}, and \texttt{explore}. Finally, to evaluate out-of-distribution generalization, we extend OGBench with a new feature for generating random maze layouts and datasets. Additional details about the environments are provided in Appendix \smartref{appendix:env}.

\subsection{Layout Generalization with Hierarchical RL}
\label{sec:hierarchical_layout_gen}

In the first experiment, we evaluate each method in a hierarchical setup. To this end, we freeze the low-level skill policy and train a high-level controller $\pi^h(z|s)$ to solve a downstream task with sparse rewards using online interaction. The high-level policy is invoked every $H$ steps, where $H$ denotes the horizon hyperparameter. During each $H$-step interval, the low-level policy takes actions conditioned on the current state and the skill selected by the high-level controller. For training the high-level policy, we employ Soft Actor-Critic (SAC) \citep{haarnoja2018sac}.

In this experiment, we compare our method against three baselines: OPAL \citep{ajay2021opal}, a strong behavioral cloning-based skill discovery method; HILP, a state-of-the-art method for distance-maximizing skill discovery; and a flat policy. For the flat baseline, we train a high-level SAC agent to output primitive actions directly rather than skills, allowing us to evaluate whether the learned skills accelerate learning on downstream tasks. (See Appendix \smartref{appendix:impl} for implementation details).

Table \smartref{table:layout_gen} reports the overall results for each environment. USB+ATR either outperforms or performs competitively with the other baselines across most environments in this setup. Although OPAL performs well on the \texttt{navigate} and \texttt{stitch} datasets in \texttt{antmaze}, it completely fails on the low-quality \texttt{explore} dataset. This strong sensitivity to dataset quality is expected from a behavioral cloning-based method, which limits its applicability in many scenarios. In contrast, HILP exhibits weaker generalization when trained on near-optimal trajectories but benefits from suboptimal, high-coverage datasets. Consequently, its performance improves significantly when provided with more diverse state coverage. The average performance on unseen layouts demonstrates that USB+ATR generalizes effectively to novel layouts, outperforming existing state-of-the-art skill discovery methods.

\subsection{Layout Generalization with Test-Time Planning}
\label{sec:planning_layout_gen}

Next, we evaluate the transferability of our method without any policy optimization in the test environment. Specifically, each method is trained on a single training layout, and the resulting policy is then transferred to an unseen layout to perform a downstream task. To provide the policy with information about the target layout, we additionally make a set of sampled states from the target layout available at test time.

We compare our approach against several baselines: GCIQL \citep{park2023hiql}, QRL \citep{wang2023qrl}, CRL \citep{eysenbach2022crl}, HIQL \citep{park2023hiql}, and HILP. GCIQL is the goal-conditioned variant of Implicit Q-Learning \citep{kostrikov2022iql}, a widely adopted method in offline RL. QRL learns a quasimetric value function by optimizing a dual objective. CRL leverages contrastive learning to estimate the value function and subsequently improve the policy. HIQL is a hierarchical approach that learns a two-level policy.

Since evaluating downstream tasks in previously unseen layouts without further policy optimization requires test-time planning, we equip each method with a planning procedure. HILP uses the planning approach introduced in its original framework. For USB+ATR, we perform graph-search-based planning over the $\phi$-embeddings of sampled states provided for each environment. For the remaining methods, we perform a similar graph-search-based planning procedure over the raw sampled states. Further details on graph construction and the planning approaches are provided in Appendix \smartref{appendix:impl}.

The results in Table \smartref{table:planning_layout_gen} indicate that USB+ATR outperforms the other methods across most environments. Although GCIQL and QRL perform well in the \texttt{pointmaze} environment, this is primarily because its 2D state space provides sufficient information for planning directly over the raw state space. In contrast, all baselines fail when evaluated in the more challenging \texttt{antmaze} environment. However, USB+ATR maintains consistently strong performance regardless of the quality of the training dataset.

\subsection{Ablation Study}
\label{sec:ablations}

\input{tables/ablation_method_component}

Our method consists of several components, and we hypothesize that each is necessary to achieve effective layout generalization. We therefore investigate the contribution of each component through an ablation study in the \texttt{antmaze-large-navigate} environment using the hierarchical evaluation setup. Specifically, we isolate each component by replacing the other components with alternative choices to assess their individual contributions. We evaluate the task and layout generalization capabilities of the following variants: HILP skills trained using our proposed ATR representation (HILP+ATR); USB directional skills trained using the universal temporal representation used in HILP (USB+UTR); USB skills trained using the representation proposed by \citet{hansen-estruch2022gcb} (USB+GCB); USB skills trained directly on the raw state representation (USB+Raw). We also compare with HILP, which we denote as HILP+UTR to distinguish the skill learning method from the temporal representation it uses.

The results in Table \smartref{table:ablation_method_component} support our hypothesis that USB is responsible for enabling layout generalization, while ATR provides the functional equivariance necessary to learn such transferable skills. Thus, the two components are complementary, with neither component alone being sufficient to achieve strong out-of-distribution layout generalization. For more details on this ablation study and additional ablations, please refer to Appendix \smartref{appendix:add_ablation}.

%% file: tables/planning_layout_generalization.tex
\begin{table}[t]
\centering
\caption{\textbf{Evaluation of transferability to new layouts using test-time planning.} The average success rates with standard errors are calculated across 5 layouts and 5 seeds for each (25 runs in total).}
\label{table:planning_layout_gen}
% \fontsize{8pt}{9pt}\selectfont
\fontsize{7.5pt}{9pt}\selectfont
% \fontsize{7pt}{8pt}\selectfont
\begin{tabular}{l l l | c c c c c c}
\toprule
\multicolumn{3}{c|}{Task} &  \multicolumn{6}{c}{Method}  \\
Environment   & Dataset  & Size  & GCIQL & QRL & CRL & HIQL & HILP & \textbf{USB+ATR} \\

\midrule

\multirow{4}{*}{\texttt{pointmaze}}
& \multirow{2}{*}{\texttt{navigate}} 
    & \texttt{medium} & $\mathbf{89.1 \smlpm 4.2}$ & $\mathbf{90.7 \smlpm 5.5}$ & ${0.0 \smlpm 0.0}$ & ${0.0 \smlpm 0.0}$ & ${0.0 \smlpm 0.0}$ & ${36.9 \smlpm 9.0}$ \\
    && \texttt{large} & ${72.2 \smlpm 9.1}$ & $\mathbf{80.0 \smlpm 8.2}$ & ${0.0 \smlpm 0.0}$ & ${0.5 \smlpm 0.5}$ & ${0.0 \smlpm 0.0}$ & ${42.0 \smlpm 9.4}$ \\
\cmidrule(lr){2-9}
& \multirow{2}{*}{\texttt{stitch}}   
    & \texttt{medium} & $\mathbf{99.3 \smlpm 0.6}$ & ${57.0 \smlpm 9.5}$ & ${0.0 \smlpm 0.0}$ & ${0.0 \smlpm 0.0}$ & ${0.0 \smlpm 0.0}$ & ${33.7 \smlpm 9.3}$ \\
    && \texttt{large} & $\mathbf{99.7 \smlpm 0.2}$ & ${80.0 \smlpm 8.2}$ & ${0.0 \smlpm 0.0}$ & ${14.2 \smlpm 6.8}$ & ${0.0 \smlpm 0.0}$ & ${52.5 \smlpm 9.8}$ \\
\midrule
\multirow{6}{*}{\texttt{antmaze}}
    & \multirow{2}{*}{\texttt{navigate}} 
        & \texttt{medium} & ${6.3 \smlpm 2.3}$ & ${0.4 \smlpm 0.4}$ & ${28.5 \smlpm 5.8}$ & ${5.0 \smlpm 1.0}$ & ${18.2 \smlpm 5.1}$ & $\mathbf{75.7 \smlpm 2.2}$ \\
        && \texttt{large} & ${0.3 \smlpm 0.3}$ & ${0.0 \smlpm 0.0}$ & ${14.0 \smlpm 3.4}$ & ${2.2 \smlpm 0.7}$ & ${1.2 \smlpm 0.7}$ & $\mathbf{66.3 \smlpm 3.9}$ \\
    \cmidrule(lr){2-9}
    & \multirow{2}{*}{\texttt{stitch}}   
        & \texttt{medium} & ${8.3 \smlpm 3.1}$ & ${0.0 \smlpm 0.0}$ & ${3.2 \smlpm 2.2}$ & ${9.1 \smlpm 3.2}$ & ${17.2 \smlpm 5.6}$ & $\mathbf{68.6 \smlpm 3.0}$ \\
        && \texttt{large} & ${9.2 \smlpm 3.0}$ & ${0.0 \smlpm 0.0}$ & ${27.8 \smlpm 6.4}$ & ${0.3 \smlpm 0.1}$ & ${3.0 \smlpm 2.1}$ & $\mathbf{54.8 \smlpm 4.3}$ \\
    \cmidrule(lr){2-9}
    & \multirow{2}{*}{\texttt{explore}}   
        & \texttt{medium} & ${57.7 \smlpm 5.5}$ & ${0.0 \smlpm 0.0}$ & ${60.4 \smlpm 6.2}$ & ${0.0 \smlpm 0.0}$ & ${20.1 \smlpm 7.3}$ & $\mathbf{69.8 \smlpm 4.1}$ \\
        && \texttt{large} & ${13.8 \smlpm 4.6}$ & ${0.0 \smlpm 0.0}$ & ${24.9 \smlpm 6.3}$ & ${0.0 \smlpm 0.0}$ & ${13.1 \smlpm 5.1}$ & $\mathbf{69.8 \smlpm 4.1}$ \\

\midrule
\multicolumn{3}{l|}{\textbf{Average}} & ${45.59}$ & ${30.81}$ & ${15.88}$ & ${3.13}$ & ${7.28}$ & $\mathbf{52.19}$ \\

\bottomrule
\end{tabular}
\end{table}

%% file: tables/ablation_method_component.tex
\begin{wraptable}[19]{r}{0.5\textwidth}
    \centering
    \caption{\textbf{Ablation study of the proposed components in the hierarchical setup.} We evaluate the task and layout generalization capabilities of each variant in the \texttt{antmaze-large-navigate} environment. Average success rates with standard errors are reported across 3 layouts/tasks, and 5 random seeds per setting (15 runs in
    total).}
    
    \label{table:ablation_method_component}
    \fontsize{9pt}{10pt}\selectfont
    \begin{tabular}{lcc}
        \toprule
        \textbf{Method} & \makecell{\textbf{Task}\\\textbf{Generalization}} & \makecell{\textbf{Layout}\\\textbf{Generalization}} \\
        \midrule
        HILP+UTR & $\mathbf{71.7 \smlpm 4.8}$  & $44.9 \smlpm 5.4$ \\
        HILP+ATR & $57.6 \smlpm 5.3$ & $25.1 \smlpm 4.3$ \\
        USB+UTR & $0.0 \smlpm 0.0$   & $0.0 \smlpm 0.0$ \\
        USB+GCB & $0.0 \smlpm 0.0$   & $0.0 \smlpm 0.0$ \\
        USB+Raw & $44.2 \smlpm 5.5$ & $28.4 \smlpm 4.1$ \\
        USB+ATR & ${66.4 \smlpm 6.4}$ & $\mathbf{83.9 \smlpm 2.9}$ \\
        \bottomrule
    \end{tabular}
\end{wraptable}

%% file: tex/6_conclusion.tex
In this work, we proposed ATR, a structured temporal representation with the functional equivariance property. We also introduced USB, which trains skills that span ATR by conditioning skill behavior on only a subset of relevant state features.
Over a set of experiments on an offline goal-conditioned benchmark, we demonstrated that our method offers strong generalization to previously unseen tasks and layouts.

\textbf{Limitations and future work.}
One limitation of our work is that we model ATR using a symmetric space. Extending the goal-conditioned representation by enforcing the triangle inequality with more general spaces, such as quasimetric architectures \citep{wang2022iqe,liu2023mrn}, is a promising direction for future research. Moreover, although our focus in this work is on deterministic environments, extending ATR to stochastic settings remains an open problem. Finally, although we only considered locomotion maze tasks in our evaluations, our method does not make any task-specific assumptions that inherently limit its applicability to this setting. Defining new benchmarks for layout generalization in other tasks, such as robotic manipulation, and evaluating existing methods on these benchmarks could be an interesting direction for future work.

%% file: tex/appendix/additional_visualizations.tex
\subsection{Visualization of ATR}
\label{appendix:viz_atr}

To illustrate how ATR enables directional skills to show monotonic behavior, we visualize a trained 32‑dimensional ATR on \texttt{\{pointmaze, antmaze\}-large-navigate} datasets and compare it with its counterpart learned by HILP. For visualization, we apply principal component analysis (PCA) to project the 32‑dimensional representations into two dimensions. PCA is used because it is a linear transformation that preserves the geometric structure of the latent space. This allows the sequence of states visited by each skill to be visualized as movements along an arbitrary vector.

\begin{figure}[t]
    \centering
    
    \begin{subfigure}[b]{0.3\textwidth}
        \centering
        \includegraphics[width=\textwidth]{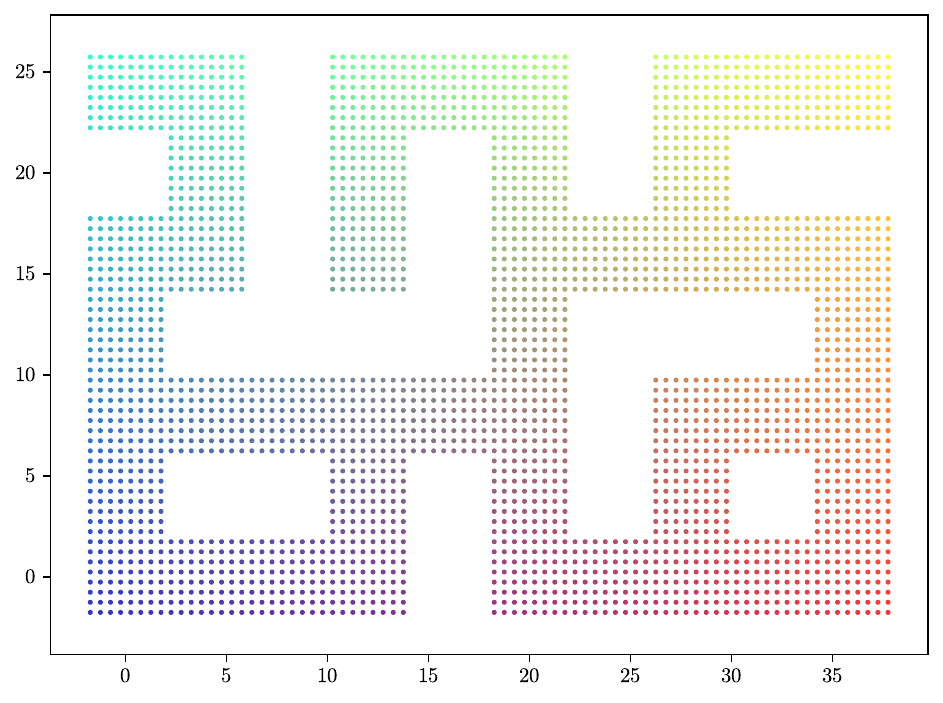}
        \caption*{Maze layout}
    \end{subfigure}
    \hfill
    \begin{subfigure}[b]{0.3\textwidth}
        \centering
        \includegraphics[width=\textwidth]{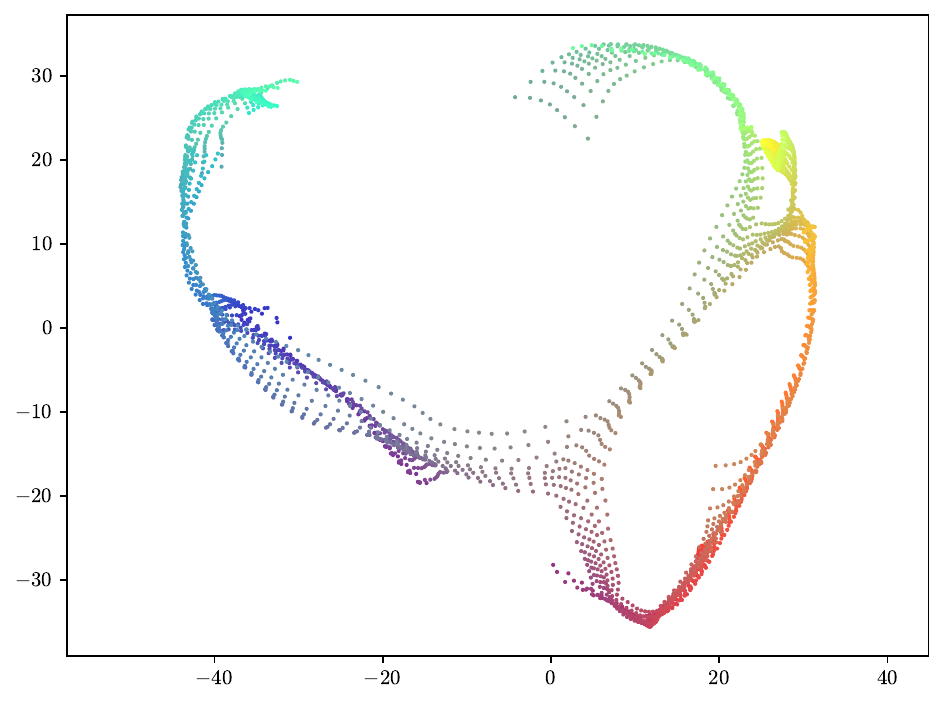}
        \caption*{HILP's temporal representation}
    \end{subfigure}
    \hfill
    \begin{subfigure}[b]{0.3\textwidth}
        \centering
        \includegraphics[width=\textwidth]{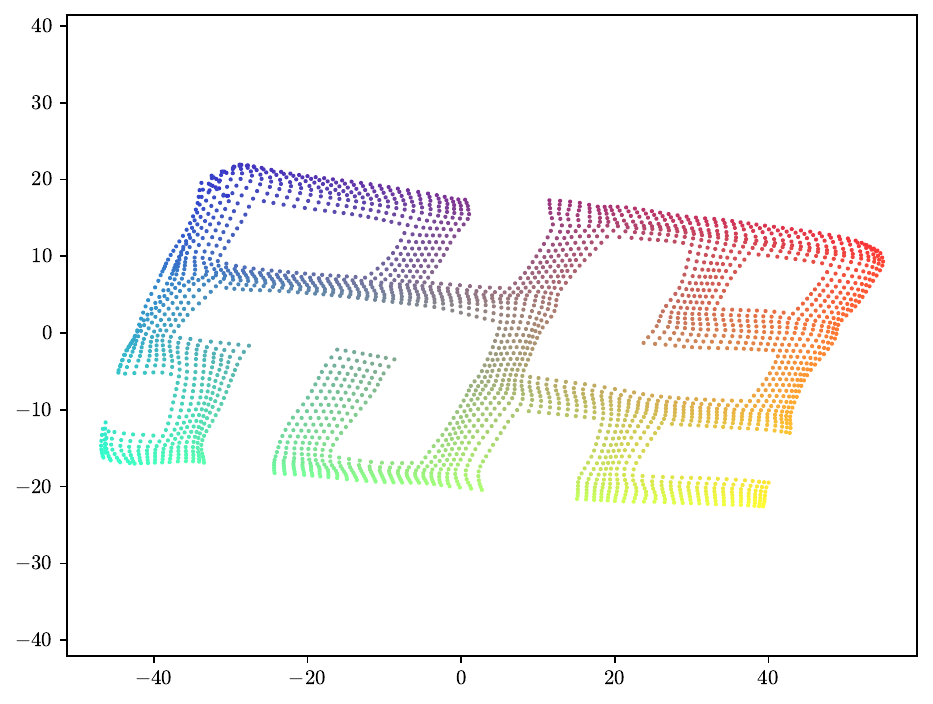}
        \caption*{ATR}
    \end{subfigure}
    
    {\texttt{antmaze-large-navigate}}
    
    \vspace{1em}
    \medskip
    
    \begin{subfigure}[b]{0.3\textwidth}
        \centering
        \includegraphics[width=\textwidth]{images/maze.pdf}
        \caption*{Maze layout}
    \end{subfigure}
    \hfill
    \begin{subfigure}[b]{0.3\textwidth}
        \centering
        \includegraphics[width=\textwidth]{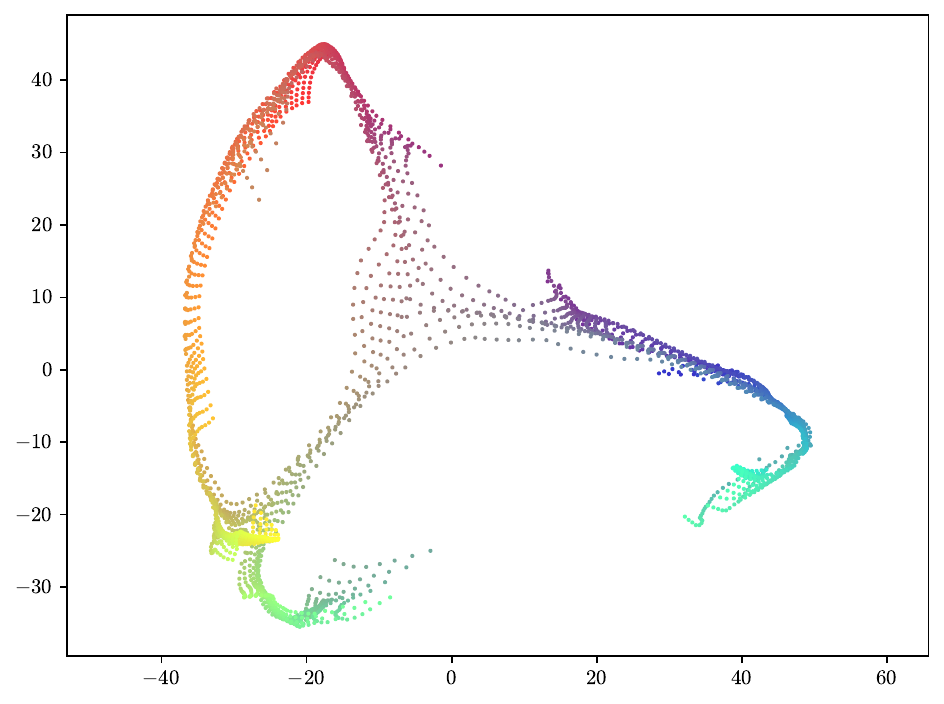}
        \caption*{HILP's temporal representation}
    \end{subfigure}
    \hfill
    \begin{subfigure}[b]{0.3\textwidth}
        \centering
        \includegraphics[width=\textwidth]{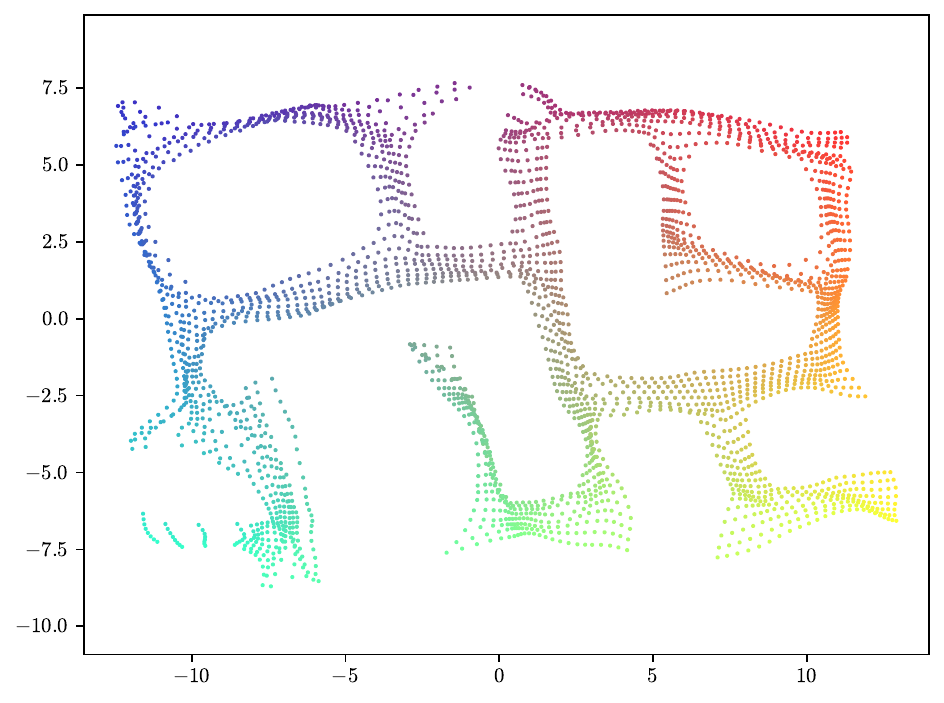}
        \caption*{ATR}
    \end{subfigure}
    
    {\texttt{pointmaze-large-navigate}}
    
    \caption{We visualize the universal temporal representation learned by HILP and ATR on a sample maze layout in two environments. ATR’s functional equivariance preserves the environment’s structural layout and aligns similar action sequences along parallel directions. In contrast, HILP distorts this structure by encoding global temporal distances and learns consistently changing behaviors.}

\label{fig:latent}
\end{figure}

Figure \smartref{fig:latent} shows that the functional equivariance property of ATR allows it to accurately capture the layout of the environment, rather than overfitting to its global temporal structure. Since similar action sequences lie on approximately parallel directions in the latent space, ATR enables USB to learn monotonic and transferable skills. In contrast, HILP learns global temporal distances between all pairs of states, which disrupts the underlying structure of the environment. As a result, it learns skills whose behaviors change consistently over time rather than following recurring action sequences.

\subsection{Visualization of $\beta$ Representation}
\label{appendix:viz_beta}

\begin{figure}[t]
    \centering
    \includegraphics[width=0.5\linewidth]{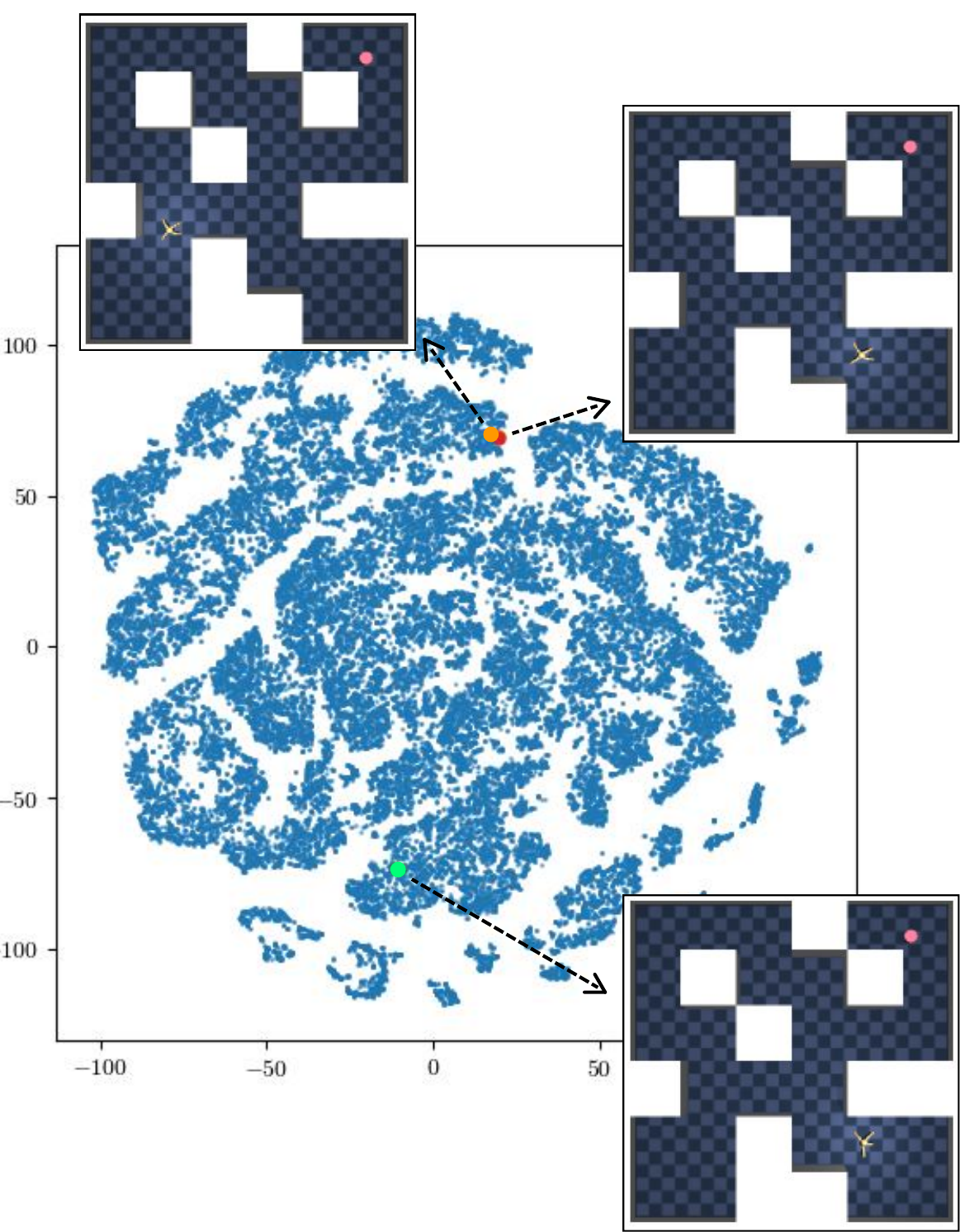}
    \caption{Visualization of the $\beta$ representation in the \texttt{antmaze-medium-navigate} environment. This embedding maps states with similar underlying dynamics to nearby representations while disregarding their temporal proximity.}
    \label{fig:beta_rep}
\end{figure}

To further evaluate the qualitative properties of our $\beta$ representation, we visualize it in the \texttt{antmaze-medium-navigate} environment. Figure \smartref{fig:beta_rep} illustrates that temporally close states can have distant embeddings in $\beta$ when their underlying dynamics differ, whereas temporally distant states with similar dynamic postures are mapped to nearby points in the representation space.

%% file: tex/appendix/additional_experiments.tex
\subsection{Qualitative Analysis of Skills using Human High-Level Control}

We qualitatively analyze the skills discovered by USB to assess their monotonicity and reusability, and to examine the extent to which the learned skills resemble those illustrated in Figure \smartref{fig:fig1b}. To this end, we first select four skills that span the first two principal components of $\phi$ in the \texttt{antmaze-large-navigate} environment. We then map these four skills, corresponding to two directions along each principal axis, to the four arrow keys. A human operator can subsequently control the ant by pressing these keys to select the corresponding skills and navigate through a previously unseen maze layout.

The human operator was able to successfully navigate the ant through the maze using only these four skills. This demonstration highlights the ability of our skill policy to generalize across layouts and execute the same skills in states that were never encountered during training. To the best of our knowledge, our method is the first to learn fully unsupervised skills that enable intuitive human control while transferring effectively to out-of-distribution maze layouts. A video of this experiment is available on the paper \href{https://sites.google.com/view/usb-atr/}{website}.

\subsection{Task Generalization}
\label{sec:task}
One natural question is whether improving transferability across different layouts comes at the cost of reduced performance on tasks within the same layout distribution. To investigate this trade-off, we also evaluate USB+ATR on in-distribution tasks using the test-time planning setup and compare its performance with that of other methods. 

The results in Table \smartref{table:planning_task_gen} show that USB+ATR maintains strong performance on standard in-distribution tasks and, on average, outperforms offline RL and skill discovery methods in the configurations for which they are specifically designed.

\input{tables/planning_task_generalization}

%% file: tables/planning_task_generalization.tex
\begin{table}[t]
\centering

\caption{\textbf{Evaluation of in-distribution task generalization using test-time planning.} The average success rates with standard errors are calculated across 5 tasks and 5 seeds for each (25 runs in total).}

\label{table:planning_task_gen}
% \fontsize{8pt}{9pt}\selectfont
\fontsize{7.5pt}{9pt}\selectfont
% \fontsize{7pt}{8pt}\selectfont
\begin{tabular}{l l l | c c c c c c}
\toprule
\multicolumn{3}{c|}{Task} &  \multicolumn{6}{c}{Method}  \\
Environment   & Dataset  & Size  & GCIQL & QRL & CRL & HIQL & HILP & \textbf{USB+ATR} \\

\midrule

\multirow{4}{*}{\texttt{pointmaze}}
& \multirow{2}{*}{\texttt{navigate}} 
    & \texttt{medium} & ${53.0 \smlpm 9.9}$ & $\mathbf{94.3 \smlpm 4.3}$ & ${16.0 \smlpm 6.6}$ & ${87.6 \smlpm 6.2}$ & ${51.9 \smlpm 9.9}$ & ${66.1 \smlpm 8.6}$ \\
    && \texttt{large} & ${30.1 \smlpm 8.3}$ & $\mathbf{84.0 \smlpm 7.5}$ & ${34.9 \smlpm 9.5}$ & ${54.9 \smlpm 8.8}$ & ${36.5 \smlpm 9.3}$ & ${70.1 \smlpm 8.1}$ \\
\cmidrule(lr){2-9}
& \multirow{2}{*}{\texttt{stitch}}   
    & \texttt{medium} & ${18.9 \smlpm 5.9}$ & ${79.0 \smlpm 7.2}$ & ${7.5 \smlpm 5.2}$ & $\mathbf{85.0 \smlpm 4.6}$ & ${22.2 \smlpm 7.9}$ & ${76.4 \smlpm 8.3}$ \\
    && \texttt{large} & ${34.2 \smlpm 8.8}$ & $\mathbf{87.4 \smlpm 6.6}$ & ${0.0 \smlpm 0.0}$ & ${10.2 \smlpm 5.8}$ & ${58.4 \smlpm 9.4}$ & ${76.0 \smlpm 8.7}$ \\
\midrule
\multirow{6}{*}{\texttt{antmaze}}
    & \multirow{2}{*}{\texttt{navigate}} 
        & \texttt{medium} & ${78.3 \smlpm 2.2}$ & ${89.1 \smlpm 1.8}$ & $\mathbf{95.5 \smlpm 0.7}$ & $\mathbf{95.0 \smlpm 0.7}$ & ${77.0 \smlpm 3.6}$ & ${83.8 \smlpm 3.7}$ \\
        && \texttt{large} & ${34.3 \smlpm 5.2}$ & ${77.9 \smlpm 2.2}$ & $\mathbf{86.4 \smlpm 2.4}$ & $\mathbf{88.4 \smlpm 3.1}$ & ${55.0 \smlpm 6.6}$ & ${64.6 \smlpm 4.8}$ \\
    \cmidrule(lr){2-9}
    & \multirow{2}{*}{\texttt{stitch}}   
        & \texttt{medium} & ${34.2 \smlpm 4.1}$ & ${68.1 \smlpm 5.4}$ & ${41.9 \smlpm 6.7}$ & $\mathbf{93.8 \smlpm 1.1}$ & ${68.2 \smlpm 5.2}$ & ${72.1 \smlpm 5.6}$ \\
        && \texttt{large} & ${8.5 \smlpm 3.7}$ & ${18.4 \smlpm 6.1}$ & ${17.6 \smlpm 5.3}$ & $\mathbf{67.3 \smlpm 6.0}$ & ${61.0 \smlpm 6.7}$ & ${49.2 \smlpm 6.1}$ \\
    \cmidrule(lr){2-9}
    & \multirow{2}{*}{\texttt{explore}}   
        & \texttt{medium} & ${11.5 \smlpm 4.0}$ & ${1.0 \smlpm 0.3}$ & ${1.9 \smlpm 0.7}$ & ${42.4 \smlpm 7.7}$ & $\mathbf{94.6 \smlpm 1.5}$ & ${84.2 \smlpm 2.4}$ \\
        && \texttt{large} & ${0.7 \smlpm 0.4}$ & ${0.0 \smlpm 0.0}$ & ${0.0 \smlpm 0.0}$ & ${1.4 \smlpm 1.1}$ & $\mathbf{35.0 \smlpm 7.1}$ & ${18.6 \smlpm 4.4}$ \\

\midrule
\multicolumn{3}{l|}{\textbf{Average}} & ${30.37}$ & ${59.92}$ & ${30.17}$ & ${62.60}$ & ${55.98}$ & $\mathbf{66.11}$ \\

\bottomrule
\end{tabular}
\end{table}

%% file: tex/appendix/additional_ablations.tex
\subsection{Ablation Study of Method Components}

In this section, we provide a deeper analysis of the results presented in Section \smartref{sec:ablations}.

The results show that HILP+ATR performs worse than HILP+UTR on unseen layouts, indicating that ATR alone does not improve layout generalization. USB+UTR fails completely because the universal temporal representation does not satisfy functional equivariance. Consequently, even for two states with identical dynamics, the same action sequence may correspond to completely different directions in UTR, making it impossible to learn a $\beta$ representation with the desired properties. USB+GCB exhibits similar behavior, as the equivariance proposed by \citet{hansen-estruch2022gcb} is defined with respect to consistency in goal-conditioned rewards rather than action consistency and therefore does not provide the functional equivariance required by USB.

In contrast, USB+Raw performs substantially better than USB+UTR and USB+GCB because the raw state representation in the OGBench maze environment partially satisfies the functional equivariance property. However, because it lacks the temporal information required for learning effective skills, it is unable to achieve performance competitive with USB+ATR.

Finally, USB+ATR significantly outperforms the other variants. Taken together, these results provide strong empirical evidence that ATR and USB are synergistic: functional equivariance enables consistent $\beta$-conditioned skills, while temporal embedding makes these skills effective, and only their combination yields the observed improvements in out-of-distribution layout generalization.

\subsection{Ablation Study of $\psi$ normalizing term}

In Equation \smartref{eq:phi_def}, we use a normalization term to maximally spread the states in the representation space. We consider several normalization options to investigate which one yields the most effective representation $\phi$.

We evaluate these options on the OGBench \texttt{antmaze-large-navigate} environment in the test-time planning setup. The results in Table \smartref{table:ablation_normalizer} indicate that $\mathbb{E}_{s \in \mathcal{S}}[\psi(s,s)]$ performs best. These results also highlight the importance of selecting an appropriate normalizer to achieve maximal state spread along the ATR.

\input{tables/ablation_normalizer}

\subsection{Ablation Study of Sensitivity to Dynamics Model}

Learning an accurate dynamics model can be challenging in many real-world applications and may result in inaccurate models. In this ablation, we investigate whether our method relies on an accurate dynamics model by evaluating its robustness to model approximation errors. Specifically, we conduct an additional ablation study in which we systematically reduce the capacity of the dynamics model and evaluate the resulting performance on the \texttt{antmaze-large-navigate} environment in the hierarchical setup.

We consider the following choices for the dynamics model:

\begin{itemize}
    \item \textbf{Small model.} A low-capacity model with hidden dimension $(64)$, resulting in approximately 10$\times$ higher MSE than the original model.
    \item \textbf{Medium model.} A medium-capacity model with hidden dimensions $(128, 128)$, resulting in approximately 4$\times$ higher MSE than the original model.
    \item \textbf{Large model.} Our original model with hidden dimensions $(512, 512, 512)$.
\end{itemize}

\input{tables/ablation_dynamics}

The results in Table \smartref{table:ablation_dynamics} show that our method achieves strong out-of-distribution generalization even when using the small dynamics model, which has approximately \textbf{10$\times$ higher MSE} than the original model.

%% file: tables/ablation_normalizer.tex
\begin{table}[tb]
\centering
\caption{\textbf{Ablation study of the $\psi$ normalization term in the test-time planning setup.} We evaluate the layout generalization of USB+ATR using different choices of the normalization term in the \texttt{antmaze-large-navigate} environment. Average success rates with standard errors are reported across 5 layouts and 5 random seeds per setting (25 runs in total).}
\label{table:ablation_normalizer}
\begin{tabular}{lc}
\hline
Normalizing Term & Success Rate \\
\hline
$\psi(s_i, s_i)$ & $0.0 \smlpm 0.0$ \\
$\psi(s_i', s_i')$ \citep{hansen-estruch2022gcb} & $0.4 \smlpm 0.2$ \\
$\mathbb{E}_{s\in\mathcal{S}}[\psi(s, s)]$ (Ours) & $\mathbf{66.3 \smlpm 3.9}$ \\
\hline
\end{tabular}
\end{table}

%% file: tables/ablation_dynamics.tex
\begin{table}[tb]
\centering
\caption{\textbf{Ablation study of sensitivity to the dynamics model in the hierarchical setup.} We evaluate the layout generalization of USB+ATR using dynamics models with different capacities in the \texttt{antmaze-large-navigate} environment. Average success rates with standard errors are reported across 3 layouts and 5 random seeds per setting (15 runs in total).}

\label{table:ablation_dynamics}
\begin{tabular}{lc}
\hline
Model Size & Success Rate \\
\hline
Small & $59.1 \smlpm 6.4$ \\
Medium & $ 69.2 \smlpm 5.5$ \\
Large (Ours) & $\mathbf{83.9 \smlpm 2.9}$ \\
\hline
\end{tabular}
\end{table}

%% file: tex/appendix/environment_details.tex
Our experiments are conducted on \texttt{antmaze} and \texttt{pointmaze} environments, along with datasets provided by OGBench. Each environment contains an agent (with 2 degrees of freedom for point and 8 degrees of freedom for ant) with the task of reaching a specified goal. We consider two environment scales: \texttt{medium} and \texttt{large}. For each environment, OGBench provides offline trajectory datasets with different qualities, which we use for pretraining skill discovery methods. The \texttt{navigate} dataset consists of trajectories collected from a noisy expert, the \texttt{stitch} dataset is composed of shorter trajectories that must be concatenated to reach distant goals, and the \texttt{explore} dataset contains extremely low-quality but high-coverage trajectories.

\begin{figure}
    \centering
    \begin{subfigure}[b]{0.3\textwidth}
        \centering
        \includegraphics[width=\textwidth]{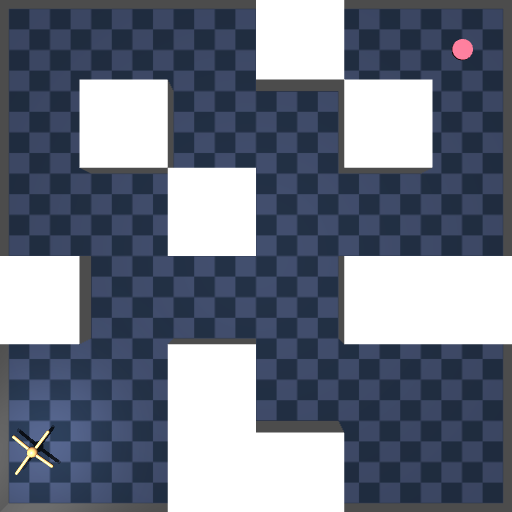}
        \caption*{seed=0 (default)}
    \end{subfigure}
    \hfill
    \begin{subfigure}[b]{0.3\textwidth}
        \centering
        \includegraphics[width=\textwidth]{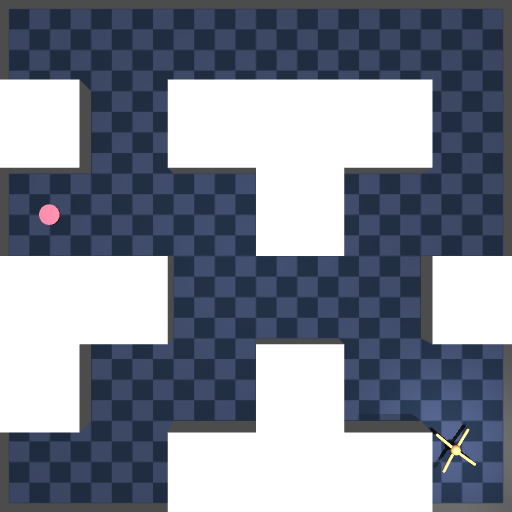}
        \caption*{seed=1}
    \end{subfigure}
    \hfill
    \begin{subfigure}[b]{0.3\textwidth}
        \centering
        \includegraphics[width=\textwidth]{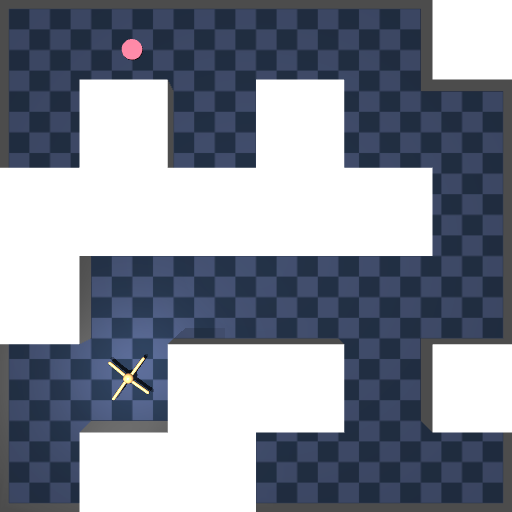}
        \caption*{seed=2}
    \end{subfigure}
    \\
    \begin{subfigure}[b]{0.3\textwidth}
        \centering
        \includegraphics[width=\textwidth]{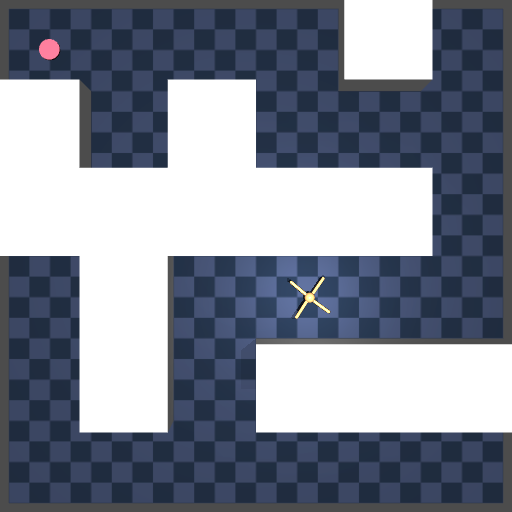}
        \caption*{seed=3}
    \end{subfigure}
    \hfill
    \begin{subfigure}[b]{0.3\textwidth}
        \centering
        \includegraphics[width=\textwidth]{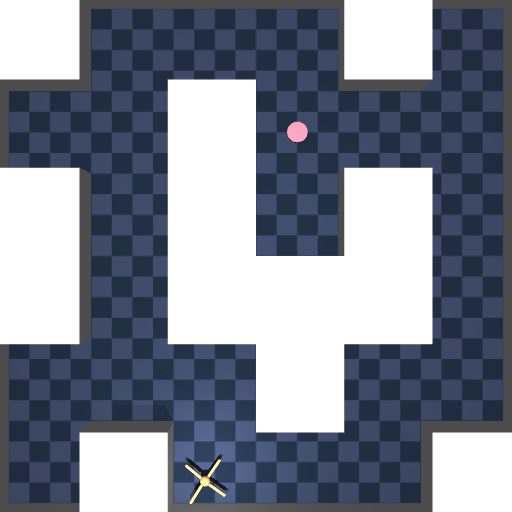}
        \caption*{seed=4}
    \end{subfigure}
    \hfill
    \begin{subfigure}[b]{0.3\textwidth}
        \centering
        \includegraphics[width=\textwidth]{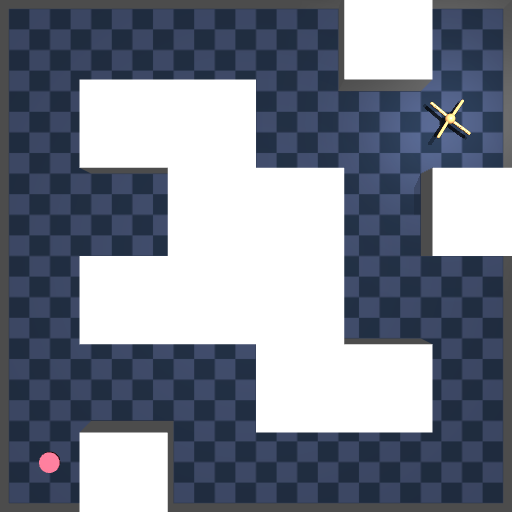}
        \caption*{seed=5}
    \end{subfigure}
    {\texttt{\{antmaze, pointmaze\}-medium}}
    
    \vspace{1em}
    \medskip

    \begin{subfigure}[b]{0.3\textwidth}
        \centering
        \includegraphics[width=\textwidth]{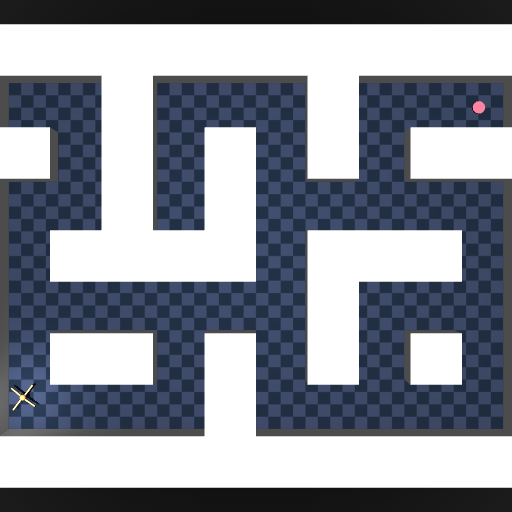}
        \caption*{seed=0 (default)}
    \end{subfigure}
    \hfill
    \begin{subfigure}[b]{0.3\textwidth}
        \centering
        \includegraphics[width=\textwidth]{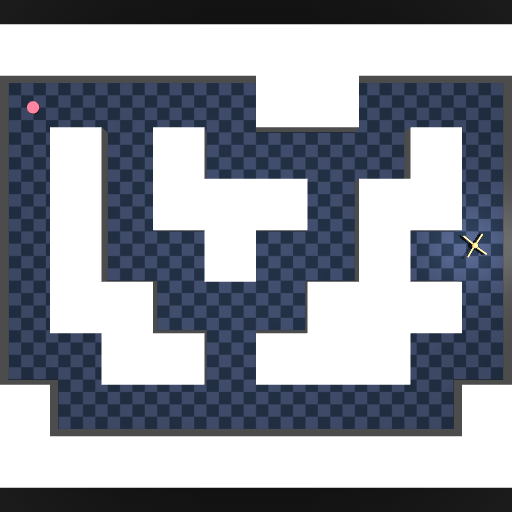}
        \caption*{seed=1}
    \end{subfigure}
    \hfill
    \begin{subfigure}[b]{0.3\textwidth}
        \centering
        \includegraphics[width=\textwidth]{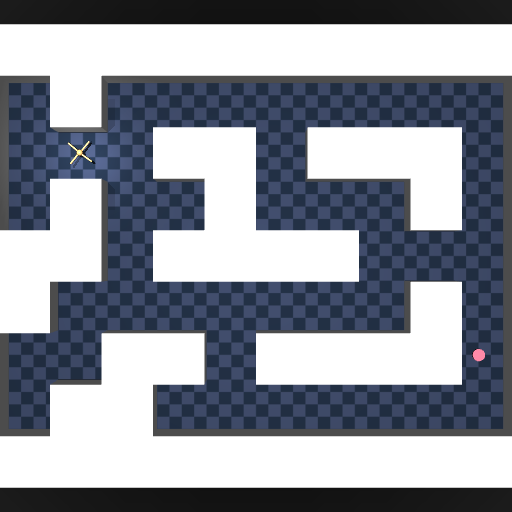}
        \caption*{seed=2}
    \end{subfigure}
    \\
    \begin{subfigure}[b]{0.3\textwidth}
        \centering
        \includegraphics[width=\textwidth]{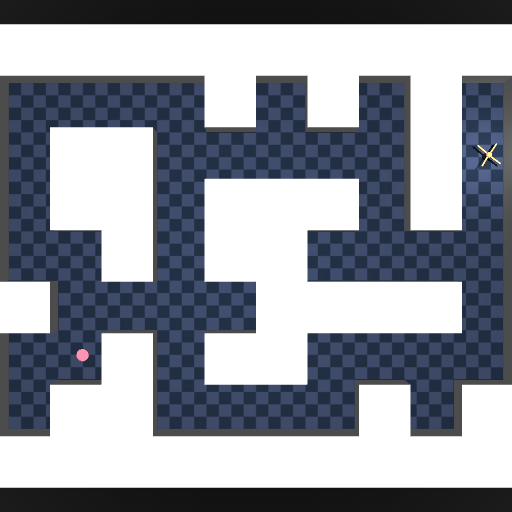}
        \caption*{seed=3}
    \end{subfigure}
    \hfill
    \begin{subfigure}[b]{0.3\textwidth}
        \centering
        \includegraphics[width=\textwidth]{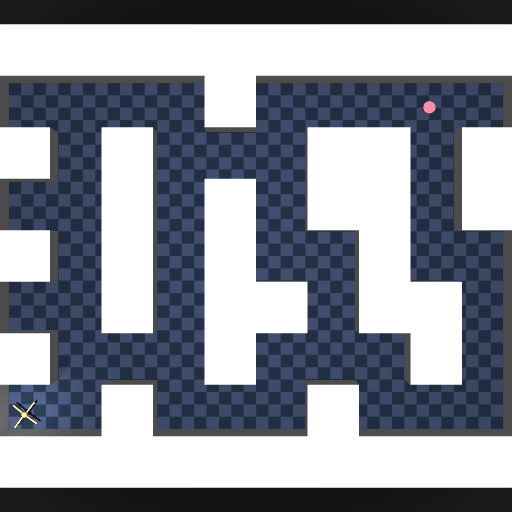}
        \caption*{seed=4}
    \end{subfigure}
    \hfill
    \begin{subfigure}[b]{0.3\textwidth}
        \centering
        \includegraphics[width=\textwidth]{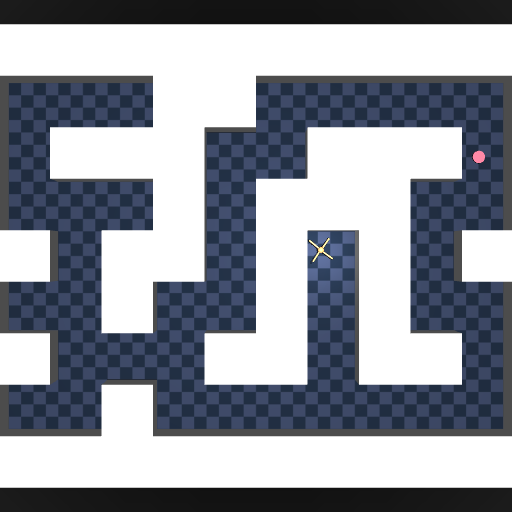}
        \caption*{seed=5}
    \end{subfigure}
    {\texttt{\{antmaze, pointmaze\}-large}}

    \caption{Randomly generated layouts in two sizes, used for evaluating the transferability of skills across different layouts}
    \label{fig:layout}
\end{figure}

To evaluate the generalization capabilities of the methods, we extended OGBench with a new feature that generates random layouts and their corresponding tasks for each environment. The generated layout with seed 0 matches the default layout and its corresponding tasks provided by OGBench, which we use to assess task-generalization performance (Appendix \smartref{sec:task}). To evaluate the transferability of learned skills across different layouts, we use the generated layouts with seeds $\{1, 2, 3,4 , 5\}$ for the test-time planning evaluation (Section \smartref{sec:planning_layout_gen}) and $\{1, 2, 3\}$ for hierarchical evaluation (Section \smartref{sec:hierarchical_layout_gen}). These layouts, along with their tasks, are shown in Figure \smartref{fig:layout}.

%% file: tex/appendix/implementation_details.tex
\textbf{Offline skill discovery.}
All our implementations are in JAX \citep{bradbury2018jax} and are based on the reference implementation codebase provided by the OGBench. For ATR, we use our own implementation of goal-conditioned bisimulation. We tuned the hyperparameter $\gamma_\psi$ by sweeping over $\{0.0, 0.5, 0.99\}$ and found that smaller values of 
$\gamma_\psi$ yield better performance on lower-quality datasets.

For USB, we use the open-source HILP implementation for the offline goal-conditioned setting, which trains skill policies using IQL \citep{kostrikov2022iql} and AWR \citep{peng2019awr}. Unlike the universal temporal representation used by HILP, ATR does not constrain the norm of the difference vector between adjacent states. Thus, to prevent training instability arising from very large or very small rewards, we found that normalizing and standardizing the directional rewards is useful in some environments. We tuned $\gamma_\beta$ via a sweep over $\{0.9, 0.95, 0.99\}$, and also performed a sweep over the dimension of $\beta$ in $\{8, 16, 32\}$. To preserve fairness across methods, we do not tune any other hyperparameters shared with HILP and instead use their suggested values. A complete report of hyperparameters used for USB+ATR is provided in Table \smartref{table:hyperparameters}.
For OPAL, since no official implementation is available, we use the implementation provided by SUPE \citep{wilcoxson2025supe}. We use the same hyperparameter values as those provided in that implementation.

\textbf{Online hierarchical RL.}
To evaluate the learned skills on downstream tasks in hierarchical setup, we use SAC \citep{haarnoja2018sac} to train a high-level policy on top of each skill discovery method through online interaction. We adopt the implementation provided by SUPE for this purpose. To determine the optimal horizon length for each method, we perform a sweep over $\{4, 8, 16\}$ and find that a horizon of 4 works best for OPAL, while 8 performs best for HILP and USB. All other hyperparameters remain unchanged compared to SUPE. We train the high-level model for up to 500K environment steps on the medium maze layout and 800K steps on the large layout. Evaluations are conducted at 300K, 400K, and 500K steps for the medium layout, and at 600K, 700K, and 800K steps for the large layout. For each evaluation step, results are averaged over 50 rollouts, and we report the mean performance across these evaluation steps.
For a fair evaluation, the high-level policy is updated every 8 environment steps in each method to keep the total number of gradient steps identical for all.

\textbf{Test-time planning.}
To evaluate the performance of the methods on unseen layouts using test-time planning, we generated a dataset for each layout and randomly sampled $N=50{,}000$ states from each dataset, which were made available to the methods at test time. For USB+ATR, we then computed the corresponding embeddings of these states in the $\phi$ space. We constructed a planning graph using a simple $K$-nearest-neighbor approach: for each state, we identified its $K$ nearest neighbors and connected them with edges weighted by their Euclidean distance. We tuned $K$ over $\{4, 8, 16\}$ and found that $K=4$ performed best. We then applied Dijkstra's algorithm to compute the shortest-path distance from each sampled state to the goal. With this preprocessing, we can dynamically identify the shortest path to the goal during execution without running Dijkstra's algorithm over the entire graph at every time step. We then select the $L$-th node along the resulting path as the subgoal, where $L$ denotes the number of lookahead steps (e.g., $L=1$ corresponds to the immediate next node), and perform the task toward this subgoal. We tune $L$ over $\{1, 2, 4, 8\}$ and find that $L=1$ performs best in \texttt{antmaze}, while $L=4$ performs best in \texttt{pointmaze}.

For all baselines other than HILP, which has its own planning procedure, we used the same planning approach but constructed the graph in the raw state space rather than in a learned representation space.

\input{tables/hyperparameters}

%% file: tables/hyperparameters.tex
\begin{table}[tb]
\centering
\caption{Hyperparameters for USB+ATR}
\label{table:hyperparameters}
\begin{tabular}{l c}
\toprule
\textbf{Hyperparameter} & \textbf{Value} \\
\midrule
\# gradient steps & $10^6$ \\
Learning rate & $0.0003$ \\
Optimizer & Adam \citep{kingma2015adam} \\
Batch size & $1024$ \\
MLP hidden dimensions & $(512, 512, 512)$ \\
Target smoothing coefficient $\tau$ & $0.005$ \\

$\psi$ discount factor $\gamma_\psi$ & $0.99$ (default), $0.0$ (\texttt{explore}) \\
$\phi$ latent dimension & $32$ \\

$\beta$ discount factor $\gamma_\beta$ & $0.99$ (default), $0.9$ (\texttt{explore}) \\
$\beta$ Latent dimension & $16$ (\texttt{antmaze}), $8$ (\texttt{pointmaze}) \\

Skills discount factor $\gamma$ & $0.99$ \\
Skills IQL expectile & $0.9$ \\
Skills AWR temperature & $10$ \\

Normalize rewards & False (\texttt{antmaze}), True (\texttt{pointmaze}) \\
\bottomrule
\end{tabular}
\end{table}